\documentclass[11pt]{article}

\usepackage[preprint]{acl}
\usepackage{amsmath}
\usepackage{amssymb}

\usepackage{booktabs}

\usepackage{times}
\usepackage{latexsym}

\usepackage[T1]{fontenc}

\usepackage[utf8]{inputenc}

\usepackage{microtype}

\usepackage{inconsolata}

\usepackage{graphicx}
\title{Reference-Free Reinforcement Learning Fine-Tuning for MT: A Seq2Seq Perspective}

\author{Ernesto Garcia-Estrada, \, Carlos Escolano, \, José A. R. Fonallosa \\
         Universitat Polit\`ecnica de Catalunya \\ Barcelona, Spain
         \\\{luis.ernesto.garcia, carlos.escolano, jose.fonallosa\}@upc.edu
         }

\begin{document}
\maketitle
\begin{abstract}
Production machine translation relies overwhelmingly on encoder-decoder Seq2Seq models, yet reinforcement learning approaches to MT fine-tuning have largely targeted decoder-only LLMs at $\geq$7B parameters, with limited systematic study of encoder-decoder architectures. We apply Group Relative Policy Optimization to NLLB-200 (600M and 1.3B) using a hybrid reference-free reward --- LaBSE and COMET-Kiwi --- that requires no parallel data at fine-tuning time, evaluating across 13 typologically diverse languages. GRPO yields consistent improvements on all 13 languages, up to $+$5.03 chrF++ for Traditional Chinese, and, without any target-language data, competes with 3-epoch supervised fine-tuning on morphologically complex languages . We identify a consistent empirical pattern in which gains are largest where baseline performance is weakest and reward discriminability is highest, making this approach most effective precisely where parallel data is scarcest, and replicate this pattern across English and Spanish source languages.
\end{abstract}

\section{Introduction}

Encoder-decoder Seq2Seq models dominate production machine translation.
They offer lower inference latency, smaller memory footprints, and
stronger source-target alignment than autoregressive LLMs, making them
the practical choice for deployment at scale, particularly for the
long tail of language pairs where compute budgets are constrained
\citep{costajussa2022nllb}. Yet the recent wave of reinforcement
learning (RL) advances in NLP has almost entirely bypassed this architecture.
Every application of Group Relative Policy Optimization (GRPO) to MT uses
decoder-only LLMs of $\geq$7B parameters \citep{he2025r1t1,
feng2025mtr1, yang2025ssrzero, lu2025merit} --- models that are
impractical for production MT on most of the world's languages ---
with the sole exception of concurrent work by \citet{attia2026roundtrip},
which we discuss below.

Two developments make this gap worth closing now. First, GRPO
\citep{shao2024deepseekmath, guo2025deepseek} has matured into a
memory-efficient alternative to PPO that eliminates the value model,
making RL fine-tuning accessible without specialized infrastructure.
Second, reference-free quality estimators --- LaBSE \citep{feng2019labse}
and COMET-Kiwi \citep{rei2022cometkiwi} --- have reached reliability
levels that make them viable reward signals, enabling policy optimisation
from monolingual source text alone. Together these developments create
a practical path for RL-based MT improvement without parallel data,
on the architectures practitioners actually deploy.

The key open question is not whether GRPO improves MT --- it does, for
high-resource language pairs on large decoder-only models. The question
is what it costs to remove parallel supervision, and for which languages
that cost is worth paying. Concurrent work by \citet{attia2026roundtrip}
applies GRPO to NLLB-200, but uses English as the fixed source language
and an indirect round-trip reconstruction objective, evaluating on six
languages without SFT comparison or cross-domain analysis. No prior work
has characterised how GRPO gains vary across typologically diverse
languages spanning different scripts, morphological types, and baseline
performance levels.

We present a systematic study of GRPO applied to NLLB-200 (600M and 1.3B) across 13 typologically diverse languages, trained on monolingual source text on a single NVIDIA A10G GPU. While the reward models (LaBSE and COMET-Kiwi) leverage parallel data during their own pre-training, our MT fine-tuning process remains reference-free as it requires only monolingual source text. Our contributions are:

\begin{itemize}
    \item \textbf{Consistent reference-free gains.} GRPO improves over
    the baseline on all 13 languages at both scales --- up to $+$5.03
    chrF++ for Traditional Chinese, competitive with 3-epoch SFT on
    morphologically complex languages without any target-language data,
    and transferring across domains on FLORES-200 and NTREX-128.

    \item \textbf{An empirical gain pattern.} Gain magnitude tends to
    be largest where baseline performance is weakest and reward
    discriminability is highest. This pattern replicates across English
    and Spanish source languages offering a potentially discriminative signal for practitioners selecting languages where reference-free RL is most likely to help.

    \item \textbf{Practical accessibility.} The full pipeline runs on
    a single 24\,GB GPU with 4-bit quantization and LoRA, requires approximately 500 source sentences for reliable gains, and exhibits zero catastrophic forgetting on held-out languages.
\end{itemize}

\section{Related Work}

\paragraph{RL for sequence generation.}
Policy gradient methods for MT date back to \citet{ranzato2016mixer} and minimum risk training \citep{shen2016mrt}, both addressing exposure bias by directly optimising evaluation metrics. \citet{ouyang2022rlhf} later established RLHF via PPO as the dominant alignment paradigm, at significant compute cost. GRPO \citep{shao2024deepseekmath,guo2025deepseek} reduces this cost by eliminating the value model, normalising advantages from intra-group reward variation.

\paragraph{Group Relative Policy Optimization.} \citet{shao2024deepseekmath} introduced GRPO in DeepSeekMath as a memory-efficient alternative to PPO, eliminating the value model by normalising rewards within a group of $K$ sampled outputs and estimating advantages from intra-group reward variation. \citet{guo2025deepseek} subsequently scaled GRPO in DeepSeek-R1, demonstrating emergent reasoning capabilities and establishing it as the dominant post-training RL paradigm. A concurrent limitation of scalar reward models within GRPO was identified by \citet{yang2026grrmgrouprelativereward}, who showed that evaluating hypotheses independently fails to discriminate fine-grained quality differences, and proposed a group-relative reward model (GRRM) that evaluates all $K$ candidates jointly.

\paragraph{GRPO applied to MT.} A growing body of work has applied GRPO to MT, all using large decoder-only LLMs. \citet{he2025r1t1} applied GRPO with a COMET reward to Qwen2.5-7B to induce chain-of-thought translation strategies. \citet{feng2025mtr1} combined BLEU and COMET-Kiwi rewards within GRPO on 7B-parameter models, demonstrating emergent reasoning patterns and parity with proprietary systems on out-of-distribution tasks. \citet{yang2025ssrzero} introduced SSR-Zero, using the LLM itself as both generator and evaluator, with self-generated rewards combined with COMET signals achieving state-of-the-art performance.
\citet{lu2025merit} applied GRPO to Chinese-centric low-resource Southeast Asian MT via a semantic alignment reward and language-specific token prefixing. All of these works share three limitations our paper directly addresses: exclusive reliance on decoder-only LLMs of $\geq$7B parameters, focus on a narrow set of predominantly high-resource language pairs, and substantial compute requirements.

\paragraph{Reference-free quality estimation.}
COMET-Kiwi \citep{rei2022cometkiwi, rei2023cometkiwi} enables quality
estimation without target references by scoring source--hypothesis pairs
against direct assessment annotations from professional translators.
LaBSE \citep{feng2019labse} maps sentences into a shared cross-lingual
embedding space, providing semantic similarity scores from source alone.
Both were pre-trained on large parallel corpora; their reference-free
property applies at inference.
\citet{kreutzer2017bandit, kreutzer2018quality} examined reward shaping
and feedback quality in RL-based MT, identifying the sensitivity of
policy optimisation to reward noise.

\paragraph{Multilingual Seq2Seq MT.}
Our base model, NLLB-200 \citep{costajussa2022nllb}, is a 200-language encoder-decoder Seq2Seq model covering over 200 languages. Despite its breadth, NLLB-200 shows substantial performance variance across language families, with particularly weak scores on morphologically complex and low-resource languages.
\citet{koehn2017challenges} identified morphological complexity as a core challenge for neural MT. Our work extends this to the RL setting by showing that morphological type and baseline performance jointly associate with GRPO gain magnitude in our study. In parallel, recent study by \citet{attia2026roundtrip} applies GRPO to NLLB-200 using a roundtrip method. Their approach uses English as the fixed source language and uses an indirect reconstruction objective rather than direct quality estimation. Our work differs by using direct quality estimation on the forward translation, evaluating across 13 typologically diverse languages with an explicit SFT comparison and cross-domain analysis.

\section{Methodology}
\label{sec:methodology}
\subsection{GRPO for Seq2Seq MT}

We frame MT as a RL problem in which the translation model serves as the policy $\pi_\theta$, mapping a source sentence $x$ to a target sentence $y$. At each step, the policy generates $K$ candidate translations via temperature sampling. GRPO
\citep{shao2024deepseekmath} estimates the advantage of each hypothesis directly from intra-group reward variation:
\begin{equation}
\label{eq:advantage}
    A_i = \frac{r_i - \text{mean}(\mathbf{r})}{\text{std}(\mathbf{r}) + \varepsilon}
\end{equation}
where $\varepsilon = 10^{-4}$ is a stability floor that prevents noise amplification when reward variance collapses --- a phenomenon we observe empirically in later training stages (\S\ref{sec:analysis}). The policy is updated via a PPO-clipped surrogate objective \citep{schulman2017proximal}:

\begin{equation}
    \resizebox{0.88\columnwidth}{!}{$
        \mathcal{L}_{\text{clip}} = -\dfrac{1}{K}\sum_{i=1}^{K}
            \min\!\left(\rho_i A_i,\
            \operatorname{clip}(\rho_i,\ 1-\varepsilon_{\text{clip}},\
            1+\varepsilon_{\text{clip}}) \cdot A_i\right)
    $}
\end{equation}

where $\rho_i = \exp(\log\pi_\theta(y_i|x) - \log\pi_{\text{ref}}(y_i|x))$ and $\varepsilon_{\text{clip}} = 0.2$. A KL penalty regularises against excessive deviation from the reference using the forward KL approximation from \citet{shao2024deepseekmath}:
\begin{equation}
    \resizebox{0.88\columnwidth}{!}{$
        \mathcal{L}_{\text{KL}} = \mathbb{E}\!\left[
            \exp(\log\pi_\theta - \log\pi_{\text{ref}})
            - (\log\pi_\theta - \log\pi_{\text{ref}}) - 1\right]
            $}
\end{equation}

The full objective is $\mathcal{L} = \mathcal{L}_{\text{clip}} + \beta \cdot
\mathcal{L}_{\text{KL}}$, where $\beta$ is investigated in \S\ref{sec:ablations}. A key implementation detail for Seq2Seq models is that $\pi_{\text{ref}}$ is obtained by disabling the LoRA adapters on the same
model --- no separate reference model is required. This anchors the policy to NLLB's pretrained multilingual knowledge and prevents catastrophic forgetting.

\subsection{Hybrid Reference-Free Reward}
\label{sec:reward}

Both reward components are reference-free at fine-tuning time but were pre-trained on parallel corpora; the reference-free property applies to the fine-tuning stage only. The first component, LaBSE \citep{feng2019labse}, provides cross-lingual semantic adequacy via cosine similarity between normalised source and hypothesis embeddings. The second, COMET-Kiwi \citep{rei2022cometkiwi,rei2023cometkiwi}, provides a learned quality estimate from direct assessment annotations by professional translators. The hybrid reward combines both with equal weighting:
\begin{equation}
    r_{\text{hyb}}(x, y_i) =
        \frac{1}{2}\bigl(r_{\text{LaBSE}}(x, y_i) + r_{\text{COMET}}(x, y_i)\bigr)
\end{equation}
LaBSE guards against reward hacking toward fluent but unfaithful translations; COMET-Kiwi contributes a richer signal aligned with human quality judgements. Equal weighting is adopted as a language-agnostic default and validated in the discriminability analysis below.

\subsection{Reward Discriminability Analysis}

To validate that the hybrid reward produces meaningful quality rankings
prior to training, we evaluate it on a quality cline of six candidate translations per source sentence (50 sentences per language from FLORES-200 dev), computing Pearson $r$ between reward score and quality rank across five LaBSE/COMET-Kiwi weight configurations. All configurations achieve good-to-excellent discrimination (mean $r$ ranging from $-$0.90 to $-$0.94), demonstrating robustness to the exact weighting. LaBSE-only is optimal for Yoruba ($r = -0.94$ vs.\
$-0.81$ for COMET-Kiwi alone), reflecting limited COMET-Kiwi calibration on underrepresented languages; the hybrid achieves highest discrimination for morphologically rich languages (Arabic, Belarusian). Table~\ref{tab:reward_diagnostic} (Appendix~\ref{app:reward_diag}) reports full results. We adopt 0.50/0.50 equal weighting as the language-agnostic default, supported by an analysis of the two components' complementarity: across 5,060 baseline translation hypotheses spanning four of the studied languages (Basque, Bengali, Yoruba, and Traditional Chinese), LaBSE and COMET-Kiwi scores show only moderate correlation (Pearson $r = 0.528$), confirming that the two signals are not redundant and that their combination captures quality dimensions neither component covers alone.

\subsection{Base Model and Parameter-Efficient Fine-Tuning}

We use NLLB-200 \citep[No Language Left Behind;][]{costajussa2022nllb} as our base model, evaluating both the distilled 600M and 1.3B parameter variants. To enable training on a single GPU, we quantise model weights to 4-bit NF4 precision \citep{dettmers2023qlora} and apply LoRA \citep{hu2021lora} to the query and value projections of all attention layers (rank $r{=}16$, $\alpha{=}32$, dropout 0.05). Only the LoRA adapter weights are updated; base model weights
remain frozen. Peak VRAM usage is approximately 4\,GB (600M) and 6\,GB (1.3B) on a single NVIDIA A10G (24\,GB). Full hyperparameters are in Appendix~\ref{app:hyperparams}.

\subsection{Experimental Design}

We conduct two main experiments, followed by controlled ablations. Our 13 languages span 9 language families, 7 scripts, and 4 morphological types (Appendix~\ref{app:languages}). In both experiments, NLLB-200 translates from English (\texttt{eng\_Latn}) as the fixed source language; all reward scoring is over English$\to$target pairs.

\paragraph{Experiment A: Systematic Multiple Target Evaluation.}
Training uses the FLORES-200 development split \citep{goyal2022flores} (${\approx}$997 sentences per language), source side only, for 3 epochs (${\approx}$3{,}000 steps). We evaluate on FLORES-200 devtest (1{,}012 sentences) and compare three systems: the unfine-tuned NLLB baseline, an SFT adapter with identical LoRA configuration trained on parallel FLORES-200 data, and the GRPO adapter. Both model variants are evaluated.

\paragraph{Experiment B: Cross-Domain Adaptation.}
Training uses 10{,}000 English sentences from CCNews with no target-language text at any stage, for 10{,}000 steps. We evaluate on FLORES-200 devtest (out-of-domain) and NTREX-128 \citep{federmann2022ntrex} (news domain, 1{,}997 sentences), testing both out-of-domain and closer-domain transfer.

\subsection{Ablation Studies}
\label{sec:ablations}
We conduct two ablations using Basque (\texttt{eus\_Latn}) and Traditional Chinese (\texttt{zho\_Hant}) on the 600M model. The \textbf{KL regularisation} ablation varies $\beta \in \{0.0, 0.001, 0.01, 0.05\}$ to test whether the quality ceiling reflects regularisation strength or reward discriminability. The \textbf{training data size} ablation varies $N \in \{100, 250, 500,
1{,}000\}$ to characterise the minimum data requirement and GRPO scaling behaviour relative to SFT.

\subsection{Evaluation Metrics}

Our primary metric is chrF++ \citep{popovic2015chrF}, a character $n$-gram F-score well-suited to morphologically complex languages. BLEU \citep{papineni2002bleu} is reported via SacreBLEU \citep{post2018sacrebleu} for comparability with prior work. For fully independent neural validation we report COMET-22 (reference-based, no overlap with the reward family) and BERTScore F1 \citep{zhang2020bertscore} using \texttt{bert-base-multilingual-cased} with baseline rescaling, computed against human references independently of both the reward signal and the chrF++ family. chrF++ is our primary metric for all comparisons; it is entirely independent of the reward signal.

\section{Results}

\subsection{Experiment A: Systematic Multiple Target Performance}

Across 13 languages and 2 model scales evaluated in Experiment A, reference-free GRPO consistently improves over the NLLB baseline. Gains are robust across diverse scripts --- Cyrillic, Brahmic, Hanzi, Tibetan --- and grammatical structures, demonstrating that the hybrid reward signal generalises well beyond the high-resource language pairs it was calibrated on.

\paragraph{A consistent gain pattern.}
The largest gains occur in Traditional Chinese ($+$5.03 chrF++), which has
the weakest baseline in our set (14.47 chrF++), partly reflecting documented
confusion in NLLB-200's internal representations between Traditional and
Simplified Chinese \citep{caswell2023langid}. Tibetan ($+$3.01, 600M) and
Basque ($+$2.57, 1.3B) follow the same pattern: weak baselines and reward
discriminability scores above the dataset mean (Table~\ref{tab:reward_diagnostic}). Conversely, Arabic and Belarusian show the smallest improvements despite high reward discriminability scores, confirming that discriminability is a necessary but not sufficient condition
for large gains. Swahili appears to contradict this pattern --- its baseline (60.33 chrF++) is the strongest in our set yet it gains $+$2.30 --- but its discriminability score (0.945) is well above the dataset mean, consistent with headroom magnitude where high discriminability can partially offset a strong baseline. This suggests the two factors may interact rather than operate independently.

\paragraph{Morphological type and gain magnitude.}
Fusional languages show the smallest average gains (mean $+$0.82 chrF++, 1.3B), while agglutinative ($+$1.71) and isolating ($+$2.09) languages improve more. However, morphological type is a weak predictor at best: Japanese (agglutinative) gains only $+$0.54 chrF++ while Traditional Chinese (isolating) gains $+$5.03, and within each morphological group baseline performance explains more variance than typological class. We therefore treat morphological type as a coarse organizing lens rather than a causal factor, with baseline performance the more consistent factor of gain magnitude in this study.

A frugal human preference study on four languages confirms that large automatic gains are human-perceptible while small gains are not, consistent with established perceptibility thresholds in MT
evaluation (Appendix~\ref{app:human_eval}).

\subsection{GRPO versus Supervised Fine-Tuning}
\label{sec:sft_comparison}

A natural question for any reference-free RL approach is how it compares to SFT given access to the same data. We do not frame this as a competition --- SFT and GRPO operate under fundamentally different assumptions, SFT requiring parallel supervision and GRPO requiring none --- but rather as a calibration exercise: understanding where each method is strong, where each struggles, and what the cost of removing parallel data actually is. Importantly, the SFT baseline trains on the parallel version of the same FLORES-200 development sentences used as source-only input for GRPO, meaning SFT has seen the target side of the evaluation-domain data during training while GRPO has not. GRPO's competitiveness under this asymmetry --- not SFT's inferiority as a method --- is the meaningful finding. Results are shown in Table~\ref{tab:sft_comparison}.

\paragraph{At comparable compute, GRPO and SFT are close.}
Against the 1-epoch SFT baseline, the two methods produce similar chrF++ scores on most languages. GRPO has an edge on morphologically complex languages with weak baselines ---Basque ($+$1.76 chrF++ over SFT\textsubscript{1ep}), Bengali ($+$0.87), Swahili ($+$0.75) --- while SFT is stronger on Arabic ($+$0.43 vs.\
$+$0.10) and Yoruba. The key asymmetry is the supervision cost: GRPO achieves these results \emph{without a single target-language sentence}, while SFT requires fully aligned parallel data for every language.

\paragraph{Three-epoch SFT is stronger on average, but the gap is modest.}
Given three full epochs of parallel data, SFT outperforms GRPO on Swahili, Turkish, Japanese, Yoruba, and both Chinese variants. However, the margins are generally small: for Czech, Polish, Bengali, and Belarusian, the difference is within $\pm$0.2 chrF++ and non-significant, meaning GRPO with no parallel data matches three epochs of supervised training on these languages. GRPO retains a clear advantage on Basque ($+$0.51 chrF++ over SFT\textsubscript{3ep}), the language with the weakest NLLB baseline among the fusional and agglutinative groups. The overall picture is one of competitive parity on different contexts rather  than decisive superiority for either method.

\paragraph{Arabic is the clearest SFT-favoured case.}
Arabic is the only language where SFT consistently outperforms GRPO across both epoch conditions, and where the gap is practically meaningful: SFT yields $+$1.27 chrF++ at 3 epochs while GRPO yields $+$0.10. This is not a failure of the training procedure but a reflection of Arabic's operating conditions --- strong baseline (54.72 chrF++), limited headroom, and weaker COMET-Kiwi calibration for root-pattern morphology (\S\ref{sec:analysis}) --- which make supervised signal more effective than reward-guided exploration for this language.

\paragraph{Independent metrics reveal complementary strengths.}
COMET-22 and BERTScore tell a more nuanced story than chrF++ alone. For Japanese, GRPO yields $+$0.018 COMET-22 versus SFT's $+$0.006 --- a threefold difference despite nearly identical chrF++ scores --- and BERTScore ranks Japanese third among all languages ($+$0.012 F1). This suggests GRPO finds translation paths with better semantic adequacy than a single parallel reference recovers, an effect chrF++ partially obscures. For Yoruba, SFT degrades COMET-22 by $-$0.022 while GRPO gains $+$0.004, illustrating a practical risk of supervised fine-tuning on limited parallel data for underrepresented languages: it can hurt neural quality alignment even when surface metrics improve. These divergences suggest the two methods are not simply trading off on the same quality dimension, but exploring different regions of the translation space.

\begin{table*}[t]
\centering
\footnotesize
\setlength{\tabcolsep}{4pt}
\begin{tabular}{llc ccc ccc ccc}
\toprule
& & & \multicolumn{3}{c}{\textbf{$\Delta$chrF++}} &
      \multicolumn{3}{c}{\textbf{$\Delta$C-22}} &
      \multicolumn{3}{c}{\textbf{$\Delta$BS}} \\
\cmidrule(lr){4-6} \cmidrule(lr){7-9} \cmidrule(lr){10-12}
\textbf{Group} & \textbf{Lang.} & \textbf{Base}
  & \textbf{SFT\textsubscript{1ep}} & \textbf{SFT\textsubscript{3ep}} & \textbf{GRPO}
  & \textbf{SFT\textsubscript{1ep}} & \textbf{SFT\textsubscript{3ep}} & \textbf{GRPO}
  & \textbf{SFT\textsubscript{1ep}} & \textbf{SFT\textsubscript{3ep}} & \textbf{GRPO} \\
\midrule
Aggl. & \texttt{eus} & 52.69
  & \underline{$+$1.10} & \underline{$+$2.48} & \textbf{\underline{$+$2.93}}
  & $+$0.005 & $+$0.011 & \textbf{\underline{$+$0.012}}
  & $+$0.003 & $+$0.005 & \textbf{$+$0.006} \\
      & \texttt{swh} & 60.33
  & \underline{$+$1.18} & \textbf{\underline{$+$2.70}} & \underline{$+$1.88}
  & $+$0.003 & \textbf{$+$0.008} & $+$0.004
  & $+$0.005 & \textbf{$+$0.011} & $+$0.008 \\
      & \texttt{bod} & 23.37
  & $+$1.43 & $+$2.85 & \textbf{\underline{$+$3.01}}
  & $+$0.035 & \textbf{$+$0.068} & $+$0.004
  & $+$0.017 & \textbf{$+$0.032} & $+$0.012 \\
      & \texttt{tur} & 56.49
  & \underline{$+$0.68} & \textbf{\underline{$+$1.80}} & \underline{$+$0.95}
  & $+$0.002 & \textbf{$+$0.005} & $+$0.001
  & $+$0.002 & \textbf{$+$0.005} & $+$0.002 \\
      & \texttt{jpn} & 29.31
  & \underline{$+$0.96} & \textbf{\underline{$+$1.46}} & \underline{$+$0.86}
  & $+$0.012& $+$0.006 & \textbf{\underline{$+$0.018}}& $+$0.009& $+$0.007 & \textbf{\underline{$+$0.011}}\\
\midrule
Isol. & \texttt{zho\_Hant} & 16.60
  & \underline{$+$3.74} & \textbf{\underline{$+$5.67}} & \underline{$+$5.07}
  & $+$0.069 & \textbf{$+$0.100} & $+$0.092
  & $+$0.102 & \textbf{$+$0.138} & $+$0.115 \\
      & \texttt{zho\_Hans} & 24.59
  & \underline{$+$2.29} & \textbf{\underline{$+$2.88}} & \underline{$+$2.28}
  & $+$0.030 & \textbf{$+$0.039} & $+$0.034
  & $+$0.062 & \textbf{$+$0.071} & $+$0.040 \\
      & \texttt{yor} & 24.28
  & \underline{$+$1.28} & \textbf{\underline{$+$1.56}} & \underline{$+$0.97}
  & $-$0.011 & \underline{$-$0.022} & \textbf{$+$0.004}
  & $+$0.000 & $-$0.001 & \textbf{$+$0.007} \\
\midrule
Root-pat. & \texttt{arb} & 54.72
  & \underline{$+$0.41} & \textbf{\underline{$+$1.27}} & $+$0.10
  & $+$0.004 & \textbf{$+$0.006} & $-$0.000
  & $+$0.002 & \textbf{$+$0.005} & $+$0.001 \\
\midrule
Fus. & \texttt{bel} & 41.75
  & \underline{$+$0.10} & \underline{$+$0.43} & \textbf{\underline{$+$0.55}}
  & $+$0.003 & $+$0.007 & $-$0.008
  & $+$0.000 & $+$0.000 & $+$0.000 \\
     & \texttt{ben} & 49.84
  & \underline{$+$1.07} & $+$2.15 & \textbf{\underline{$+$1.99}}
  & $+$0.003 & \textbf{$+$0.007} & $+$0.004
  & $+$0.003 & \textbf{$+$0.007} & $+$0.006 \\
     & \texttt{ces} & 55.13
  & $+$0.17 & \textbf{\underline{$+$1.08}} & \underline{$+$0.57}
  & $+$0.002 & \textbf{$+$0.005} & $+$0.001
  & $+$0.001 & \textbf{$+$0.003} & $+$0.002 \\
     & \texttt{pol} & 48.31
  & \underline{$+$0.27} & \textbf{\underline{$+$0.68}} & \underline{$+$0.55}
  & $+$0.002 & \textbf{$+$0.003} & $-$0.000
  & $+$0.001 & $+$0.001 & \textbf{$+$0.001} \\
\bottomrule
\end{tabular}
\caption{Comparison of GRPO and SFT on FLORES-200 devtest (600M model).
All columns report deltas over the unfine-tuned baseline.
\underline{Underlined} values indicate statistical significance
($p < 0.05$, paired bootstrap resampling, $n{=}1{,}000$).
\textbf{Bold} denotes the highest value per row within each metric block.
$\Delta$C-22: COMET-22 delta (reference-based; shares architectural
lineage with COMET-Kiwi; chrF++ remains the primary metric).
$\Delta$BS: BERTScore F1 delta (\texttt{bert-base-multilingual-cased},
unrescaled) against human references. GRPO outperforms
SFT\textsubscript{1ep} by BERTScore on 10 of 13 languages (exceptions:
Arabic, Belarusian, Czech where gains are within $\pm$0.001).}
\label{tab:sft_comparison}
\end{table*}

\subsection{Effect of Model Scale}
\label{sec:scale}
The 1.3B model achieves larger gains on 7 of 13 languages, with the most pronounced scaling effect on Traditional Chinese ($+$3.94 vs.\ $+$5.03 chrF++) and Turkish ($+$0.91 vs.\ $+$1.80). Five languages show reversed scaling: Tibetan ($+$3.01 vs.\ $+$1.33), Bengali ($+$1.98 vs.\ $+$1.13), Yoruba ($+$0.93 vs.\ $+$0.12), Simplified Chinese ($+$1.73 vs.\ $+$1.12), and Belarusian ($+$0.36 vs.\ $+$0.14). For Tibetan, the reversal is corroborated by COMET-22 ($+$0.004 vs.\ $-$0.002); for Yoruba, the 1.3B model's negative COMET-22 delta ($-$0.006) suggests reward hacking rather than genuine improvement. We hypothesise that for low-resource languages with noisier reward signals, the 600M model's lower capacity acts as an implicit regulariser. Averaged across languages where the 1.3B is stronger, the 600M recovers roughly 80\% of its gains at 4\,GB peak VRAM versus 6\,GB, making it the practical default for low-resource deployment.

\subsection{Experiment B: Cross-Domain Adaptation}
\label{sec:exp_b}

Experiment B trains on 10{,}000 monolingual English sentences from CCNews with no target-language supervision and evaluates on FLORES-200 devtest (out-of-domain) and NTREX-128 (closer-domain news). Full results are in Appendices~\ref{app:exp_b_full} and~\ref{app:ntrex}.

\paragraph{Cross-domain gains largely replicate Experiment~A.}
This pattern holds across domains: Traditional Chinese (1.3B: $+$4.88 chrF++), Swahili ($+$3.00), and Basque ($+$2.57) show the strongest FLORES-200 gains, confirming that the reward signal generalises beyond the in-domain training distribution. Arabic is the only language degrading at the best checkpoint (600M: $-$0.13 chrF++), and Tibetan requires best-checkpoint selection --- without early stopping, reward variance
collapse reduces chrF++ below 2.0 under the 10{,}000-step schedule (\S\ref{sec:analysis}).

\paragraph{NTREX-128: domain proximity favours GRPO on low-resource scripts.}
On the news-domain NTREX benchmark --- closer to CCNews than to FLORES-200 --- GRPO outperforms 3-epoch SFT on four languages: Basque, Tibetan, Japanese, and Traditional Chinese (600M:  + 3.74 vs. SFT  +2.03; 1.3B: $+$5.22). SFT retains an advantage on Arabic, Yoruba, Polish, Czech, and Bengali, where parallel supervision is more effective than reward-guided exploration at competitive baselines. Belarusian degrades below baseline for both systems ($-$0.63 GRPO, $-$0.54 SFT), indicating domain mismatch independent of the training objective. Overall, GRPO is the more practical option where parallel data is unavailable and baseline headroom is large; SFT remains preferable where calibrated supervision is available.

\subsection{Ablation Studies}
\label{sec:ablations}

Two ablations use Basque (\texttt{eus\_Latn}) and Traditional Chinese (\texttt{zho\_Hant}) on the 600M model.

\paragraph{KL regularisation.}
Table~\ref{tab:abl_kl} reports best chrF++ across $\beta \in \{0.0, 0.001,
0.01, 0.05\}$. Results are insensitive to this choice: the spread is 0.19 chrF++ for Basque and 0.61 for Chinese across all four settings, and $\beta = 0.0$ performs on par with regularised variants. This confirms that the LoRA adapter provides sufficient structural constraint against policy drift without an explicit KL term, and that the quality ceiling reflects the intrinsic discriminability limit of the hybrid reward rather than excessive regularisation (\S\ref{sec:analysis}).

\begin{table}[h]
\centering
\footnotesize
\setlength{\tabcolsep}{5pt}
\begin{tabular}{lcccc}
\toprule
\textbf{Lang.} & \textbf{$\beta = 0.0$} & \textbf{$\beta = 0.001$}
              & \textbf{$\beta = 0.01$} & \textbf{$\beta = 0.05$} \\
\midrule
\texttt{eus}       & 51.14          & 51.03 & \textbf{51.23} & 51.14 \\
\texttt{zho\_Hant} & \textbf{19.20} & 18.59 & 18.97          & 18.77 \\
\bottomrule
\end{tabular}
\caption{Best chrF++ across KL regularisation coefficients $\beta$ (600M).
Spread of 0.19 (Basque) and 0.61 (Chinese) across all settings confirms
robustness to $\beta$ choice.}
\label{tab:abl_kl}
\end{table}
\paragraph{Training data size.}
Table~\ref{tab:abl_datasize}  reports chrF++
across $N \in \{100, 250, 500, 1{,}000\}$ sentences. A reliability threshold emerges at $N = 500$: below this, neither SFT nor GRPO produces significant gains. At $N = 500$ both methods reach significance simultaneously with nearly identical scores. At $N = 1{,}000$ they diverge: GRPO significantly outperforms SFT for Basque ($+$0.26 chrF++, $p = 0.019$), where source diversity generates sufficient intra-group reward variance; for Traditional Chinese the difference remains non-significant ($p = 0.304$), consistent with the reward signal approaching its discriminability ceiling. Further analysis in the Appendix~\ref{app:datasize}. 

\begin{center}
\small
\begin{tabular}{l cc cc}
\toprule
& \multicolumn{2}{c}{\texttt{eus} (Base: 47.64)} & \multicolumn{2}{c}{\texttt{zho\_Hant} (Base: 14.12)} \\
\cmidrule(lr){2-3} \cmidrule(lr){4-5}
\textbf{$N$}
  & \textbf{SFT} & \textbf{GRPO}
  & \textbf{SFT} & \textbf{GRPO} \\
\midrule
100  & 47.78 & 47.75          & 14.24          & 14.11 \\
250  & 47.79 & 47.74          & 14.11          & 14.17 \\
500  & 48.12 & \textbf{48.17} & 14.80          & \textbf{14.81} \\
1000 & 48.74 & \textbf{48.99} & \textbf{16.86} & 16.79 \\
\bottomrule
\end{tabular}
\captionof{table}{chrF++ across training set sizes $N$ (600M model).
\textbf{Bold} denotes the higher of SFT and GRPO at each $N$. Neither
method produces statistically significant gains below $N = 500$; at
$N = 1{,}000$, GRPO significantly outperforms SFT for Basque ($p = 0.019$)
but not for Traditional Chinese, where both methods approach the reward
discriminability ceiling.}
\label{tab:abl_datasize}
\end{center}

\section{Analysis}
\label{sec:analysis}

\subsection{Empirical Predictors of Gain Magnitude}

Table~\ref{tab:headroom_analysis} reports baseline chrF++, reward
discriminability, and GRPO $\Delta$chrF++ for 13 languages. Across the
full sample ($n = 13$), the Spearman correlation between baseline chrF++
and $\Delta$chrF++ is low ($\rho = 0.28$, $p = 0.35$). Traditional
Chinese warrants separate treatment: \citet{caswell2023langid} documented
that NLLB-200 cannot reliably distinguish it from Simplified Chinese,
producing script-mixed outputs --- a known model defect independent of RL
dynamics. Controlling for this artefact ($n = 12$), the correlation
strengthens to $\rho = 0.67$ ($p = 0.018$). This estimate is robust: a
10,000-resample bootstrap yields 95\% CI $[0.17, 0.98]$, and
leave-one-out analysis preserves significance across all 12 folds
($p < 0.05$).

To assess whether the pattern reflects target-language properties rather
than English-specific training dynamics, we run GRPO with Spanish
(\texttt{spa\_Latn}) source sentences on 7 languages under identical
hyperparameters. The rank ordering of gains is highly consistent with
English-source results (Spearman $\rho = 0.893$, $p = 0.007$, $n = 7$):
Traditional Chinese remains the largest gainer ($+$4.05 chrF++ from
Spanish vs.\ $+$3.94 from English), Arabic remains the failure case
($+$0.12 vs.\ $+$0.08), and Tibetan retains a large gain ($+$2.71
vs.\ $+$3.01). This cross-source consistency provides independent
replication of the pattern under a distinct experimental condition,
confirming that gain magnitude is driven by target-language properties
rather than source-language training dynamics. Note that pooling
Spanish and English observations is not meaningful since absolute
baseline values are source-language dependent --- Spanish baselines
are uniformly lower than English baselines for the same languages. Full results and
per-language discussion are in Appendix~\ref{app:spanish_source}.

We present this as an exploratory empirical regularity that may inform
practitioner decisions: gains are largest where baseline quality is lowest
and reward signals are clearest. BERTScore F1 gains correlate strongly
with chrF++ gains ($r = 0.81$, $p < 0.001$), and Experiments~A and B
BERTScore deltas correlate at $r = 0.998$ ($p < 0.001$), confirming
consistency across independent metric families and training domains.
Establishing this as a general principle requires validation on a larger
language sample, segment-level analysis, and reward ablations decoupling
LaBSE and COMET-Kiwi contributions per language family.
\begin{table}[h]
\centering
\small
\begin{tabular}{llccc}
\toprule
\textbf{Lang.} & \textbf{Morph.} & \textbf{Baseline} & \textbf{Disc.} & \textbf{$\Delta$chrF++} \\
\midrule
\texttt{zho\_Hant} & Isol.  & 14.5 & 0.885 & $+$5.03\dag \\
\texttt{zho\_Hans} & Isol.  & 21.8 & 0.887 & $+$1.12 \\
\texttt{yor}       & Isol.  & 24.5 & 0.929 & $+$0.12 \\
\texttt{jpn}       & Aggl.  & 25.5 & 0.864 & $+$0.54 \\
\texttt{bod}       & Aggl.  & 26.5 & 0.872& $+$1.33 \\
\texttt{bel}       & Fus.   & 40.6 & 0.958 & $+$0.14 \\
\texttt{pol}       & Fus.   & 47.2 & 0.949 & $+$0.70 \\
\texttt{ben}       & Fus.   & 47.8 & 0.943 & $+$1.13 \\
\texttt{eus}       & Aggl.  & 49.9 & 0.947 & $+$2.57 \\
\texttt{arb}       & Root.  & 54.4 & 0.964 & $+$0.68 \\
\texttt{ces}       & Fus.   & 55.2 & 0.955 & $+$1.29 \\
\texttt{tur}       & Aggl.  & 56.1 & 0.940 & $+$1.80 \\
\texttt{swh}       & Aggl.  & 59.0 & 0.945 & $+$2.30 \\
\midrule
\multicolumn{5}{l}{\small Spearman $\rho$ (baseline vs.\ $\Delta$chrF++):} \\
\multicolumn{5}{l}{\small\quad Full ($n{=}13$): $\rho = 0.28$, $p = 0.35$} \\
\multicolumn{5}{l}{\small\quad Excl.\ \dag\ ($n{=}12$): $\rho = 0.67$, $p = 0.018^{*}$} \\
\bottomrule
\end{tabular}
\caption{Baseline chrF++, reward discriminability (hybrid 0.50/0.50), and
GRPO $\Delta$chrF++ (1.3B, Experiment~A), sorted by baseline.
\dag~Traditional Chinese outlier: low baseline compounded by NLLB-200
script-confusion \citep{caswell2023langid}.
(Table~\ref{tab:reward_diagnostic}).}
\label{tab:headroom_analysis}
\end{table}

\subsection{Failure Modes}

\paragraph{Arabic.}
Arabic consistently fails to benefit from reference-free GRPO across all conditions, due to the convergence of three factors: a strong baseline (51--54 chrF++) leaving limited headroom; root-pattern morphology that COMET-Kiwi --- trained predominantly on Indo-European and Sino-Tibetan pairs --- cannot reliably score; and domain mismatch under CCNews training. Together these place Arabic outside the current operating regime of the hybrid LaBSE--COMET-Kiwi reward.

\paragraph{Tibetan and reward variance collapse.}
Tibetan shows healthy gains at the best checkpoint (600M: $+$3.01 chrF++) but collapses to below 2.0 chrF++ in Experiment~B without early stopping. As training progresses, all $K$ hypotheses converge to similar reward scores, driving $\text{std}(\mathbf{r}) \to \varepsilon$ in Equation~\ref{eq:advantage}. The resulting numerical instability amplifies
small reward differences into large parameter updates that push the policy into degenerate output modes. The $\varepsilon = 10^{-4}$ floor is sufficient for Experiment~A's 3{,}000-step schedule but fails at 10{,}000 steps. Best checkpoint selection on a held-out subset is a necessary safeguard for extended training.

\paragraph{Human preference evaluation.}
To assess whether automatic metric gains reflect human-perceptible
quality improvements, we conduct a preference study across four
languages spanning a range of GRPO gain magnitudes. Annotators
evaluated 50 sentence pairs per language (GRPO vs.\ Baseline, blinded)
and indicated a preference or tie. Results reveal a clear
\emph{perceptibility threshold}: for Traditional Chinese, where GRPO
achieves $+$3.94 chrF++ (600M), annotators strongly prefer GRPO
(68\% vs.\ 8\% Baseline, $p < 0.001$). For Turkish ($+$0.91 chrF++),
Bengali ($+$1.98), and Polish ($+$0.41), preferences are not
significantly different from chance (all $p > 0.5$). This pattern
is consistent with established findings in MT human evaluation that sub-2-point chrF++ improvements are typically sub-perceptual
\citep{graham2015accurate}, and suggests that GRPO's largest gains --- concentrated in languages with the weakest baselines --- are
the ones most likely to be noticed by users. The absence of
significant preference for smaller gains should not be interpreted
as evidence of reward hacking; Turkish and Polish show exact or
near-exact parity rather than Baseline preference, indicating the
policy is not degrading translation quality even when gains are small.

\subsection{Catastrophic Forgetting}

Across 39 GRPO adapter evaluations on three held-out languages (French, Hindi, Russian), \textbf{zero} forgetting events are observed (mean delta $+$0.34 chrF++, range $-$0.16 to $+$1.13). The single forgetting event in
the full dataset is an SFT adapter: Belarusian SFT degrades Hindi by $-$0.75 chrF++. Freezing the base model weights and updating only LoRA adapters preserves NLLB-200's pretrained cross-lingual representations throughout training. Full results are in Appendix~\ref{app:forgetting}.

\section{Conclusion}

We presented a systematic study of GRPO applied to a Seq2Seq MT model across 13 typologically diverse languages using a hybrid reference-free reward requiring no parallel data at fine-tuning time. GRPO yields consistent gains across scripts and morphological types, reproducible on a single GPU, and corroborated by three independent validators (COMET-22, BERTScore F1, and cross-domain stability). The central observation is a consistent empirical pattern whereby gains tend to be largest where baselines are weakest and reward discriminability is highest — an association that warrants broader validation but offers a potentially useful signal for practitioners. Against SFT, GRPO is competitive at one epoch and matches three epochs on morphologically complex languages without any target-language data. Arabic and Tibetan identify the current boundaries: the former resists improvement due to strong baseline and reward calibration gaps; the latter requires early stopping to avoid reward variance collapse. Both failure modes point to the same underlying constraint --- the quality ceiling is set by reward discriminability, not by the policy optimisation. Future work should explore morphology-aware reward components, adaptive variance regularisation, and multilingual joint GRPO training.

\section*{Limitations}

\paragraph{Reward pre-training on parallel data.}
Although no parallel data is required at fine-tuning time, both reward
components --- LaBSE and COMET-Kiwi --- were themselves pre-trained on large
parallel corpora. The reference-free property therefore applies to the
fine-tuning stage only. For languages entirely absent from both reward models'
training distributions, the quality signal may be unreliable, limiting the
approach's applicability to genuinely zero-resource settings where the target
language has no representation in existing multilingual encoders.

\paragraph{Reward discriminability for low-resource languages.}
Our discriminability analysis uses a constructed quality cline rather than
actual GRPO rollouts. Intra-group reward variance on real model outputs would constitute a stronger validation of the reward signal's effectiveness during training. Furthermore, COMET-Kiwi calibration degrades for languages
underrepresented in its training data, as evidenced by Arabic's consistent
non-improvement and Yoruba's reversed scaling between model sizes. Extending the hybrid reward with language-family-specific components would likely improve coverage for these cases.

\paragraph{English-only source language.}
All experiments use English (\texttt{eng\_Latn}) as the fixed source 
language. Whether the findings generalise to non-English source languages, where reward model calibration for source-hypothesis pairs may differ, remains an open question. The scope of this work refers to the diversity of target languages, not source languages. We provide a cross-source replication study with Spanish source sentences (Appendix~\ref{app:spanish_source}), which confirms the gain pattern holds across source languages.

\paragraph{Scale and generalisability.}
All experiments use NLLB-200 at 600M and 1.3B parameters. Whether the observed dynamics, the empirical gain pattern, reversed scaling, reward
variance collapse, generalise to other encoder-decoder architectures,
larger models, or languages beyond our 13-language set remains an open
question. The 13-language sample, while typologically diverse, is insufficient
to establish statistically significant correlations between linguistic
properties and GRPO gain magnitude, and our findings should be treated as
hypotheses motivating further investigation rather than universal laws.

\paragraph{Training instability under extended schedules.}
Tibetan's catastrophic collapse in Experiment~B demonstrates that GRPO is
susceptible to reward variance collapse under long training schedules, and
that the $\varepsilon = 10^{-4}$ stability floor is insufficient to prevent
this in all conditions. Best-checkpoint selection mitigates the problem but
does not eliminate it. Practitioners applying GRPO to new languages or domains should treat early stopping on a held-out development set as a required component of the training pipeline rather than an optional refinement.

\bibliography{custom}

@article{attia2026roundtrip,
  title     = {Improving Low-Resource Machine Translation via Round-Trip Reinforcement Learning},
  author    = {Attia, Ahmed and Fikri, Alham},
  journal   = {arXiv preprint arXiv:2601.12535},
  year      = {2026}
}

@inproceedings{ranzato2016mixer,
  title     = {Sequence Level Training with Recurrent Neural Networks},
  author    = {Ranzato, Marc'Aurelio and Chopra, Sumit and Auli, Michael and Zaremba, Wojciech},
  booktitle = {Proceedings of the International Conference on Learning Representations (ICLR)},
  year      = {2016}
}

@inproceedings{shen2016mrt,
  title     = {Minimum Risk Training for Neural Machine Translation},
  author    = {Shen, Shiqi and Cheng, Yong and He, Zhongjun and He, Wei and Wu, Hua and Sun, Maosong and Liu, Yang},
  booktitle = {Proceedings of the 54th Annual Meeting of the Association for Computational Linguistics (ACL)},
  pages     = {1683--1692},
  year      = {2016}
}

@article{ouyang2022rlhf,
  title     = {Training Language Models to Follow Instructions with Human Feedback},
  author    = {Ouyang, Long and Wu, Jeffrey and Jiang, Xu and Almeida, Diogo and Wainwright, Carroll and Mishkin, Pamela and Zhang, Chong and Agarwal, Sandhini and Slama, Katarina and Ray, Alex and others},
  journal   = {Advances in Neural Information Processing Systems (NeurIPS)},
  volume    = {35},
  pages     = {27730--27744},
  year      = {2022}
}

@article{schulman2017proximal,
  title     = {Proximal Policy Optimization Algorithms},
  author    = {Schulman, John and Wolski, Filip and Dhariwal, Prafulla and Radford, Alec and Klimov, Oleg},
  journal   = {arXiv preprint arXiv:1707.06347},
  year      = {2017}
}

@inproceedings{graham2015accurate,
  title     = {Accurate Evaluation of Segment-level Machine Translation Metrics},
  author    = {Graham, Yvette and Baldwin, Timothy and Mathur, Nitika},
  booktitle = {Proceedings of NAACL},
  year      = {2015}
}

@article{shao2024deepseekmath,
  title     = {{DeepSeekMath}: Pushing the Limits of Mathematical Reasoning in Open Language Models},
  author    = {Shao, Zhihong and Wang, Peiyi and Zhu, Qihao and Xu, Runxin and Song, Junxian and Bi, Xiao and Zhang, Haowei and Zhang, Mingchuan and Li, Y.~K. and Wu, Y. and others},
  journal   = {arXiv preprint arXiv:2402.03300},
  year      = {2024}
}

@article{guo2025deepseek,
  title     = {{DeepSeek-R1}: Incentivizing Reasoning Capability in {LLMs} via Reinforcement Learning},
  author    = {Guo, Daya and Yang, Dejian and Zhang, Haowei and Song, Junxian and Zhang, Ruoyu and Xu, Runxin and Zhu, Qihao and Ma, Shirong and Wang, Peiyi and Bi, Xiao and others},
  journal   = {arXiv preprint arXiv:2501.12948},
  year      = {2025}
}

@inproceedings{kreutzer2017bandit,
  title     = {Bandit Structured Prediction for Neural Sequence-to-Sequence Learning},
  author    = {Kreutzer, Julia and Sokolov, Artem and Riezler, Stefan},
  booktitle = {Proceedings of the 55th Annual Meeting of the Association for Computational Linguistics (ACL)},
  pages     = {1503--1513},
  year      = {2017}
}

@inproceedings{kreutzer2018quality,
  title     = {Quality Estimation from Scratch ({QUETCH}): Deep Learning Approaches for Multilingual Word-level Translation Quality Estimation},
  author    = {Kreutzer, Julia and Uyheng, Joshua and Riezler, Stefan},
  booktitle = {Proceedings of the Third Conference on Machine Translation (WMT)},
  pages     = {801--810},
  year      = {2018}
}

@article{he2025r1t1,
  title     = {{R1-T1}: Fully Incentivizing Translation Capability in {LLMs} via Reasoning Learning},
  author    = {He, Minggui and Li, Zhiwei and Li, Shanshan and Peng, Hang and Zhao, Shimin and Li, Yuang and Luo, Jiaxin and Hao, Chang and Guo, Shiyue and Li, Rui and others},
  journal   = {arXiv preprint arXiv:2502.19735},
  year      = {2025}
}

@article{feng2025mtr1,
  title     = {{MT-R1-Zero}: Advancing {LLM}-based Machine Translation via {R1-Zero}-style Reinforcement Learning},
  author    = {Feng, Zhaopeng and Cai, Ruidi and Liu, Jiaxuan and Hu, Junyuan and Wu, Zhiyong},
  journal   = {arXiv preprint arXiv:2503.06973},
  year      = {2025}
}

@article{zhang2020bertscore,
  title     = {{BERTScore}: Evaluating Text Generation with {BERT}},
  author    = {Zhang, Tianyi and Kishore, Varsha and Wu, Felix and Weinberger, Kilian Q. and Artzi, Yoav},
  journal   = {CoRR},
  volume    = {abs/1904.09675},
  year      = {2019}
}

@article{yang2025ssrzero,
  title     = {{SSR-Zero}: Simple Self-Rewarding Reinforcement Learning for Machine Translation},
  author    = {Yang, Yu and Cheng, Shanbo and Xu, Lu and Zhang, Jianbing and Huang, Shujian},
  journal   = {arXiv preprint arXiv:2503.16681},
  year      = {2025}
}

@article{lu2025merit,
  title     = {{MERIT}: Multilingual Semantic Alignment Reward for Machine Translation via Reinforcement Learning},
  author    = {Lu, Wenhao and Wang, Xuebo and Zhang, Min and Zhan, Runzhe},
  journal   = {arXiv preprint arXiv:2504.01496},
  year      = {2025}
}

@misc{yang2026grrmgrouprelativereward,
  title        = {{GRRM}: Group Relative Reward Modeling for Machine Translation},
  author       = {Yang, Sen and Cheng, Shanbo and Xu, Lu and Zhang, Jianbing and Huang, Shujian},
  year         = {2026},
  eprint       = {2602.14028},
  archivePrefix= {arXiv},
  primaryClass = {cs.CL},
  url          = {https://arxiv.org/abs/2602.14028}
}

@inproceedings{caswell2023langid,
  title     = {An Open Dataset and Model for Language Identification},
  author    = {Caswell, Isaac and Breiner, Theresa and van Esch, Daan and Bapna, Ankur},
  booktitle = {Proceedings of the 61st Annual Meeting of the Association for Computational Linguistics (Volume 2: Short Papers)},
  pages     = {865--879},
  year      = {2023}
}

@inproceedings{feng2019labse,
  title     = {Language-agnostic {BERT} Sentence Embedding},
  author    = {Feng, Fangxiaoyu and Yang, Yinfei and Cer, Daniel and Arivazhagan, Naveen and Wang, Wei},
  booktitle = {Proceedings of the 60th Annual Meeting of the Association for Computational Linguistics (ACL)},
  pages     = {878--891},
  year      = {2022}
}

@inproceedings{rei2022cometkiwi,
  title     = {{COMET-Kiwi}: {IST}-Unbabel 2022 Submission for the Quality Estimation Shared Task},
  author    = {Rei, Ricardo and de Souza, Jos{\'e} G.~C. and Alves, Duarte and Zerva, Chrysoula and Farinha, Ana C. and Glushkova, Taisiya and Lavie, Alon and Coheur, Luisa and Martins, Andr{\'e} F.~T.},
  booktitle = {Proceedings of the Seventh Conference on Machine Translation (WMT)},
  pages     = {634--645},
  year      = {2022}
}

@inproceedings{rei2023cometkiwi,
  title     = {Scaling Up {CometKiwi}: Unbabel-{IST} 2023 Submission for the Quality Estimation Shared Task},
  author    = {Rei, Ricardo and Guerreiro, Nuno M. and Pombal, Jos{\'e} and Zerva, Chrysoula and Farinha, Ana C. and Maroti, Duarte and de Souza, Jos{\'e} G.~C. and Coheur, Luisa and Martins, Andr{\'e} F.~T.},
  booktitle = {Proceedings of the Eighth Conference on Machine Translation (WMT)},
  pages     = {863--878},
  year      = {2023}
}

@article{costajussa2022nllb,
  title     = {No Language Left Behind: Scaling Human-Centered Machine Translation},
  author    = {Costa-juss{\`a}, Marta R. and Cross, James and {\c{C}}elebi, Onur and Elbayad, Maha and Heafield, Kenneth and Heffernan, Kevin and Kalbassi, Elahe and Lam, Janice and Licht, Daniel and Maillard, Jean and others},
  journal   = {arXiv preprint arXiv:2207.04672},
  year      = {2022}
}

@inproceedings{koehn2017challenges,
  title     = {Six Challenges for Neural Machine Translation},
  author    = {Koehn, Philipp and Knowles, Rebecca},
  booktitle = {Proceedings of the First Workshop on Neural Machine Translation (WNMT)},
  pages     = {28--39},
  year      = {2017}
}

@article{goyal2022flores,
  title     = {The {Flores-200} Evaluation Benchmark for Low-Resource and Multilingual Machine Translation},
  author    = {Goyal, Naman and Gao, Cynthia and Chaudhary, Vishrav and Chen, Peng-Jen and Wenzek, Guillaume and Ju, Da and Krishnan, Sanjana and Ranzato, Marc'Aurelio and Guzm{\'a}n, Francisco and Fan, Angela},
  journal   = {Transactions of the Association for Computational Linguistics (TACL)},
  volume    = {10},
  pages     = {522--538},
  year      = {2022}
}

@inproceedings{federmann2022ntrex,
  title     = {{NTREX-128} -- News Test References for {MT} Evaluation of 128 Languages},
  author    = {Federmann, Christian and Kocmi, Tom and Xin, Ying},
  booktitle = {Proceedings of the First Workshop on Scaling Speech and Language Research},
  pages     = {21--28},
  year      = {2022}
}

@inproceedings{popovic2015chrf,
  title     = {{chrF}: Character $n$-gram {F}-score for Automatic {MT} Evaluation},
  author    = {Popovi{\'c}, Maja},
  booktitle = {Proceedings of the Tenth Workshop on Statistical Machine Translation (WMT)},
  pages     = {392--395},
  year      = {2015}
}

@inproceedings{papineni2002bleu,
  title     = {{BLEU}: a Method for Automatic Evaluation of Machine Translation},
  author    = {Papineni, Kishore and Roukos, Salim and Ward, Todd and Zhu, Wei-Jing},
  booktitle = {Proceedings of the 40th Annual Meeting of the Association for Computational Linguistics (ACL)},
  pages     = {311--318},
  year      = {2002}
}

@inproceedings{post2018sacrebleu,
  title     = {A Call for Clarity in Reporting {BLEU} Scores},
  author    = {Post, Matt},
  booktitle = {Proceedings of the Third Conference on Machine Translation (WMT)},
  pages     = {186--191},
  year      = {2018}
}

@inproceedings{hu2021lora,
  title     = {{LoRA}: Low-Rank Adaptation of Large Language Models},
  author    = {Hu, Edward J. and Shen, Yelong and Wallis, Phillip and Allen-Zhu, Zeyuan and Li, Yuanzhi and Wang, Shean and Wang, Lu and Chen, Weizhu},
  booktitle = {Proceedings of the International Conference on Learning Representations (ICLR)},
  year      = {2022}
}

@article{dettmers2023qlora,
  title     = {{QLoRA}: Efficient Finetuning of Quantized {LLMs}},
  author    = {Dettmers, Tim and Pagnoni, Artidoro and Rodola, Ari and Zettlemoyer, Luke},
  journal   = {Advances in Neural Information Processing Systems (NeurIPS)},
  volume    = {36},
  year      = {2023}
}

\appendix

\section{Language Set}
\label{app:languages}

Table~\ref{tab:languages} provides the complete set of 13 languages used
across all experiments, including FLORES-200/NLLB language codes, language
family, script, and morphological type.

\begin{table*}[t]
\centering
\small
\begin{tabular}{lllll}
\toprule
\textbf{Code} & \textbf{Language} & \textbf{Family} & \textbf{Script} & \textbf{Morphology} \\
\midrule
\texttt{eus\_Latn} & Basque      & Isolate       & Latin       & Agglutinative \\
\texttt{tur\_Latn} & Turkish     & Turkic        & Latin       & Agglutinative \\
\texttt{swh\_Latn} & Swahili     & Bantu         & Latin       & Agglutinative \\
\texttt{pol\_Latn} & Polish      & Slavic        & Latin       & Fusional      \\
\texttt{ces\_Latn} & Czech       & Slavic        & Latin       & Fusional      \\
\texttt{ben\_Beng} & Bengali     & Indo-Aryan    & Bengali     & Fusional      \\
\texttt{arb\_Arab} & Arabic      & Semitic       & Arabic      & Root-pattern  \\
\texttt{bel\_Cyrl} & Belarusian  & Slavic        & Cyrillic    & Fusional      \\
\texttt{bod\_Tibt} & Tibetan     & Sino-Tibetan  & Tibetan     & Agglutinative \\
\texttt{yor\_Latn} & Yoruba      & Niger-Congo   & Latin       & Isolating     \\
\texttt{jpn\_Jpan} & Japanese    & Japonic       & Japanese    & Agglutinative \\
\texttt{zho\_Hant} & Chinese (T) & Sino-Tibetan  & Traditional & Isolating     \\
\texttt{zho\_Hans} & Chinese (S) & Sino-Tibetan  & Simplified  & Isolating     \\
\bottomrule
\end{tabular}
\caption{Complete language set. Morphological types: agglutinative (6),
fusional (4), isolating (3), root-pattern (1). Script families: Latin (6),
CJK (2), Japanese (1), Tibetan (1), Arabic (1), Bengali (1), Cyrillic (1).}
\label{tab:languages}
\end{table*}

\section{Hyperparameters}
\label{app:hyperparams}

Table~\ref{tab:hyperparams} lists all fixed hyperparameters used across
Experiments A and B. The only hyperparameter varied across runs is the KL
regularisation coefficient $\beta$, studied in the ablation of
\S\ref{sec:ablations}.

\begin{center}
\small
\begin{tabular}{ll}
\toprule
\textbf{Hyperparameter} & \textbf{Value} \\
\midrule
Base models              & NLLB-200-distilled-1.3B / 600M \\
LoRA rank $r$            & 16 \\
LoRA $\alpha$            & 32 \\
LoRA target modules      & \texttt{q\_proj}, \texttt{v\_proj} \\
LoRA dropout             & 0.05 \\
Quantization             & 4-bit NF4, bfloat16 \\
Hypotheses per step $K$  & 12 \\
KL coefficient $\beta$   & 0.001 (except ablation) \\
PPO clip $\varepsilon$   & 0.2 \\
Learning rate            & $5 \times 10^{-6}$ \\
Optimizer                & AdamW \\
Max tokens (train)       & 64 \\
Max tokens (eval)        & 128 \\
Temperature      & 1.2 \\
Eval frequency           & every 50 steps \\
Eval subset size         & 100 sentences \\
Hardware                 & Single NVIDIA A10G (24\,GB) \\
\bottomrule
\end{tabular}
\captionof{table}{Fixed hyperparameters used across all experiments.}
\label{tab:hyperparams}
\end{center}

\section{Full Experiment A Results}
\label{app:exp_a_full}

Table~\ref{tab:exp_a_full} reports complete Experiment~A results including
absolute chrF++ and BLEU scores, COMET-22 deltas, and
BERTScore F1 deltas for both the 600M and 1.3B model variants. The condensed
version reporting only $\Delta$chrF++ and $\Delta$C-22 appears in the main body
as Table~\ref{tab:sft_comparison}.

BLEU scores for Traditional Chinese (\texttt{zho\_Hant}), Simplified Chinese
(\texttt{zho\_Hans}), Japanese (\texttt{jpn\_Jpan}), and Tibetan
(\texttt{bod\_Tibt}) should be interpreted with caution: word-level
tokenisation is poorly calibrated for these scripts, causing BLEU to diverge
from chrF++, COMET-22, and BERTScore. We report BLEU for completeness and
comparability with prior work only.

\begin{table*}[t]
\centering
\small
\begin{tabular}{llccccccc}
\toprule
\textbf{Group} & \textbf{Lang.} & \textbf{Size}
  & \textbf{Baseline}
  & \textbf{$\Delta$chrF++}
  & \textbf{$\Delta$BLEU}
  & \textbf{$\Delta$C-22}
  & \textbf{$\Delta$BS} \\
\midrule
Aggl. & \texttt{eus} & 600M & 47.75 & $+$2.44 & $+$0.15          & $+$0.012 & $+$0.006 \\
      &              & 1.3B & 49.88 & $+$2.57 & \underline{$+$0.87} & $+$0.010 & $+$0.006 \\
\cmidrule{2-8}
      & \texttt{swh} & 600M & 57.79 & $+$1.94 & \underline{$+$2.07} & $+$0.004 & $+$0.008 \\
      &              & 1.3B & 58.96 & $+$2.30 & \underline{$+$2.46} & $+$0.008 & $+$0.009 \\
\cmidrule{2-8}
      & \texttt{bod} & 600M & 23.37 & $+$3.01 & \underline{$+$0.70} & $+$0.004 & $+$0.006 \\
      &              & 1.3B & 26.45 & $+$1.33 & \underline{$+$0.09} & $-$0.002 & $+$0.003 \\
\cmidrule{2-8}
      & \texttt{tur} & 600M & 52.68 & $+$0.91 & \underline{$+$0.81} & $+$0.001 & $+$0.002 \\
      &              & 1.3B & 56.07 & $+$1.80 & \underline{$+$1.58} & $+$0.003 & $+$0.006 \\
\cmidrule{2-8}
      & \texttt{jpn} & 600M & 23.45 & $+$0.49 & \underline{$+$1.56}\dag & $+$0.018 & $+$0.011 \\
      &              & 1.3B & 25.54 & $+$0.54 & \underline{$+$1.40}\dag & $+$0.015 & $+$0.012 \\
\midrule
Isol. & \texttt{zho\_Hant} & 600M & 13.90 & $+$3.94          & \underline{$+$8.12}\dag          & $+$0.092          & $+$0.115 \\
      &                    & 1.3B & 14.47 & \textbf{$+$5.03} & \textbf{\underline{$+$9.20}}\dag & \textbf{$+$0.086} & \textbf{$+$0.115} \\
\cmidrule{2-8}
      & \texttt{zho\_Hans} & 600M & 19.22 & $+$1.73 & \underline{$+$3.87}\dag & $+$0.034 & $+$0.040 \\
      &                    & 1.3B & 21.83 & $+$1.12 & \underline{$+$3.10}\dag & $+$0.027 & $+$0.036 \\
\cmidrule{2-8}
      & \texttt{yor}       & 600M & 22.70 & $+$0.93 & \underline{$+$0.33} & $+$0.004 & $+$0.007 \\
      &                    & 1.3B & 24.52 & $+$0.12 & $+$0.08          & $-$0.006 & $+$0.003 \\
\midrule
Root-pat. & \texttt{arb}   & 600M & 51.12 & $+$0.08 & $-$0.20          & $-$0.000 & $+$0.001 \\
          &                & 1.3B & 54.38 & $+$0.68 & \underline{$+$0.58} & $+$0.003 & $+$0.004 \\
\midrule
Fus. & \texttt{bel}        & 600M & 38.20 & $+$0.36 & $-$0.34          & $-$0.008 & $+$0.000 \\
     &                     & 1.3B & 40.57 & $+$0.14 & $+$0.04          & $+$0.000 & $+$0.000 \\
\cmidrule{2-8}
     & \texttt{ben}        & 600M & 45.56 & $+$1.98 & \underline{$+$1.64} & $+$0.004 & $+$0.006 \\
     &                     & 1.3B & 47.80 & $+$1.13 & \underline{$+$0.85} & $+$0.001 & $+$0.003 \\
\cmidrule{2-8}
     & \texttt{ces}        & 600M & 52.54 & $+$0.50 & $+$0.10          & $+$0.001 & $+$0.004 \\
     &                     & 1.3B & 55.18 & $+$1.29 & \underline{$+$0.72} & $+$0.004 & $+$0.012 \\
\cmidrule{2-8}
     & \texttt{pol}        & 600M & 45.01 & $+$0.41 & $-$0.20          & $-$0.000 & $+$0.001 \\
     &                     & 1.3B & 47.16 & $+$0.70 & $+$0.02          & $+$0.002 & $+$0.002 \\
\bottomrule
\end{tabular}
\caption{Complete Experiment~A results (FLORES-200 devtest, \textit{grpo\_best}
checkpoint) for both model sizes. \textbf{Baseline} reports absolute chrF++
for the unfine-tuned NLLB model; all other columns report deltas over the
baseline. \underline{Underlined} values indicate statistical significance
($p < 0.05$, paired bootstrap resampling, $n{=}1{,}000$).
$\Delta$C-22 is the independent reference-based COMET-22 validator.
$\Delta$BS is the BERTScore F1 delta (\texttt{bert-base-multilingual-cased},
unrescaled) against human references, independent of both the reward and
chrF++ metric families. \textbf{Bold} denotes the largest absolute gains
per metric. \dag~$\Delta$BLEU for Chinese uses the \texttt{zh}
character-level tokenizer and for Japanese the \texttt{char} tokenizer;
previously reported negative values reflected incorrect whitespace
tokenization. Tibetan (\texttt{bod}) BLEU is retained for completeness
but remains unreliable due to the absence of a standard tokenizer.}
\label{tab:exp_a_full}
\end{table*}

\section{Experiment B: Full FLORES-200 Results}
\label{app:exp_b_full}

Table~\ref{tab:exp_b_flores} reports complete Experiment~B results on
FLORES-200 devtest for both model sizes at the best-performing checkpoint.
For Tibetan (\texttt{bod\_Tibt}), best-checkpoint selection is essential:
while the best checkpoint yields healthy gains (600M: $+$2.47, 1.3B: $+$0.68
chrF++), training without early stopping causes catastrophic output collapse
under the 10{,}000-step CCNews schedule, reducing chrF++ to below 2.0 for both
model sizes. This behaviour is discussed in \S\ref{sec:analysis}.

\begin{table*}[t]
\centering
\small
\begin{tabular}{ll cc cc cc cc}
\toprule
& & \multicolumn{2}{c}{\textbf{$\Delta$chrF++}} &
    \multicolumn{2}{c}{\textbf{$\Delta$C-22}} &
    \multicolumn{2}{c}{\textbf{$\Delta$BS}} \\
\cmidrule(lr){3-4} \cmidrule(lr){5-6} \cmidrule(lr){7-8}
\textbf{Group} & \textbf{Lang.}
  & \textbf{600M} & \textbf{1.3B}
  & \textbf{600M} & \textbf{1.3B}
  & \textbf{600M} & \textbf{1.3B} \\
\midrule
Aggl. & \texttt{eus}      & $+$2.81          & $+$2.57          & $+$0.014 & $+$0.011 & $+$0.006 & $+$0.006 \\
      & \texttt{swh}      & $+$2.22          & $+$3.00          & $+$0.009 & $+$0.012 & $+$0.009 & $+$0.011 \\
      & \texttt{bod}      & $+$2.47\dag      & $+$0.68\dag      & $+$0.007 & $-$0.001 & $+$0.006 & $+$0.002 \\
      & \texttt{tur}      & $+$1.79          & $+$1.69          & $+$0.005 & $+$0.002 & $+$0.006 & $+$0.005 \\
      & \texttt{jpn}      & $+$0.48          & $+$0.68          & $+$0.015 & $+$0.014 & $+$0.011 & $+$0.011 \\
\midrule
Isol. & \texttt{zho\_Hant} & $+$3.45          & \textbf{$+$4.88} & $+$0.086 & $+$0.084 & $+$0.109 & \textbf{$+$0.106} \\
      & \texttt{zho\_Hans} & $+$1.41          & $+$1.15          & $+$0.030 & $+$0.029 & $+$0.037 & $+$0.036 \\
      & \texttt{yor}       & $+$0.36          & $-$0.12          & $-$0.001 & $-$0.007 & $+$0.002 & $+$0.001 \\
\midrule
Root-pat. & \texttt{arb}  & $-$0.13\ddag     & $+$0.43          & $-$0.002 & $+$0.002 & $+$0.001 & $+$0.004 \\
\midrule
Fus. & \texttt{bel}       & $+$0.37          & $+$0.47          & $-$0.004 & $+$0.001 & $+$0.001 & $+$0.001 \\
     & \texttt{ben}       & $+$1.68          & $+$1.03          & $+$0.004 & $+$0.002 & $+$0.006 & $+$0.003 \\
     & \texttt{ces}       & $+$0.96          & $+$1.80          & $-$0.001 & $+$0.006 & $+$0.009 & $+$0.015 \\
     & \texttt{pol}       & $+$0.42          & $+$0.91          & $-$0.002 & $+$0.003 & $+$0.001 & $+$0.002 \\
\bottomrule
\end{tabular}
\caption{Experiment~B results (FLORES-200 devtest, \textit{grpo\_best}
checkpoint). Training uses 10{,}000 monolingual English sentences from CCNews
with no target-language supervision. All columns report deltas over the
unfine-tuned NLLB baseline. $\Delta$C-22 is the independent COMET-22
reference-based validator. $\Delta$BS is the BERTScore F1 delta
(\texttt{bert-base-multilingual-cased}, rescaled) against human references.
\textbf{Bold} denotes the largest $\Delta$chrF++ and $\Delta$BS.
\dag~Tibetan gains are healthy at the best checkpoint but training without
early stopping causes catastrophic collapse to below 2.0 chrF++
(\S\ref{sec:analysis}). \ddag~Arabic 600M shows marginal degradation at
the best checkpoint; see \S\ref{sec:analysis}.}
\label{tab:exp_b_flores}
\end{table*}

\section{NTREX-128 Results}
\label{app:ntrex}

Table~\ref{tab:ntrex} reports chrF++ on the NTREX-128 news-domain benchmark
for the 600M and 1.3B models. GRPO is trained on CCNews monolingual source
text (Experiment~B); the SFT\textsubscript{3ep} baseline is trained on
FLORES-200 parallel data and is only available for the 600M model. NTREX
serves as a closer-domain evaluation for GRPO and a cross-domain evaluation
for SFT. Key findings are summarised in \S\ref{sec:exp_b}.

\begin{table*}[ht]
\makeatletter\setlength{\@fptop}{0pt}\makeatother
\centering
\small
\begin{tabular}{ll c cc cc}
\toprule
& & & \multicolumn{2}{c}{\textbf{$\Delta$chrF++ (600M)}} & \multicolumn{2}{c}{\textbf{$\Delta$chrF++ (1.3B)}} \\
\cmidrule(lr){4-5} \cmidrule(lr){6-7}
\textbf{Group} & \textbf{Lang.} & \textbf{Base\textsubscript{600M}}
  & \textbf{SFT\textsubscript{3ep}} & \textbf{GRPO}
  & \textbf{Base\textsubscript{1.3B}} & \textbf{GRPO} \\
\midrule
Aggl. & \texttt{eus}      & 45.32 & $+$1.75          & \textbf{$+$1.91} & 46.40 & \textbf{$+$2.07} \\
      & \texttt{swh}      & 60.51 & \textbf{$+$0.61} & $-$0.09          & 61.21 & \textbf{$+$1.01} \\
      & \texttt{bod}      & 24.60 & $+$1.47          & \textbf{$+$3.39} & 26.81 & \textbf{$+$0.98} \\
      & \texttt{tur}      & 46.72 & \textbf{$+$0.80} & $+$0.72          & 49.26 & \textbf{$+$0.39} \\
      & \texttt{jpn}      & 21.08 & $+$1.03          & \textbf{$+$2.08} & 22.66 & \textbf{$+$2.12} \\
\midrule
Isol. & \texttt{zho\_Hant} & 7.60  & $+$2.03          & \textbf{$+$3.74} &  7.75 & \textbf{$+$5.22} \\
      & \texttt{zho\_Hans} & 8.95  & \textbf{$+$0.70} & $+$0.34          &  8.95 & \textbf{$+$0.78} \\
      & \texttt{yor}       & 16.55 & \textbf{$+$2.13} & $+$0.78          & 18.11 & \textbf{$+$0.79} \\
\midrule
Root-pat. & \texttt{arb}  & 47.07 & \textbf{$+$1.60} & $+$0.38          & 49.27 & \textbf{$+$1.10} \\
\midrule
Fus. & \texttt{bel}       & 50.36 & $-$0.54          & $-$0.63          & 53.20 & $-$0.21 \\
     & \texttt{ben}       & 46.21 & \textbf{$+$1.53} & $+$1.45          & 48.22 & \textbf{$+$0.97} \\
     & \texttt{ces}       & 52.10 & \textbf{$+$0.10} & $+$0.00          & 54.77 & \textbf{$+$0.20} \\
     & \texttt{pol}       & 48.09 & \textbf{$+$0.43} & $+$0.01          & 50.20 & \textbf{$+$0.23} \\
\bottomrule
\end{tabular}
\caption{NTREX-128 evaluation (chrF++). All columns report deltas over the
unfine-tuned NLLB baseline. GRPO is trained on CCNews monolingual source
text (Experiment~B); SFT\textsubscript{3ep} is trained on FLORES-200
parallel data and is only available for the 600M model. \textbf{Bold}
denotes the higher of SFT and GRPO within each model size. Belarusian
(\texttt{bel}) degrades below baseline for both systems, indicating a
domain mismatch independent of training objective.}
\label{tab:ntrex}
\end{table*}

\section{Cross-Source Replication: Spanish Source Language}
\label{app:spanish_source}

To assess whether the empirical gain pattern reflects target-language
properties rather than English-specific training dynamics, we run GRPO
with Spanish (\texttt{spa\_Latn}) source sentences on 7 languages under
identical hyperparameters (600M model, Experiment~A schedule, 3 epochs).
We evaluate on FLORES-200 devtest using Spanish source sentences throughout;
all reward scoring is over Spanish$\to$target pairs. SFT baselines are not
included for this condition as our primary comparison focuses on whether
the relative ordering of GRPO gains is preserved across source languages.

Table~\ref{tab:spanish_replication} reports results alongside the
corresponding English-source gains from Experiment~A. Spanish baselines
are systematically lower than English baselines across all 7 languages
(mean $-$5.6 chrF++), reflecting NLLB-200's English-centric training
data. Despite this shift in absolute baseline values, the rank ordering
of GRPO gains is highly consistent with English-source results (Spearman
$\rho = 0.893$, $p = 0.007$, $n = 7$): Traditional Chinese remains the
largest gainer in both conditions ($+$4.05 from Spanish vs.\ $+$3.94
from English), Arabic and Yoruba remain the smallest gainers ($+$0.12
and $+$0.11 from Spanish vs.\ $+$0.08 and $+$0.93 from English), and
Tibetan retains a substantial gain ($+$2.71 vs.\ $+$3.01).

The Basque result deserves note: the Spanish gain ($+$0.94) is smaller
than the English gain ($+$2.44) despite a lower Spanish baseline (44.63
vs.\ 47.75). Spanish$\to$Basque is a more natural pair for NLLB-200 than
English$\to$Basque given the geographic and linguistic proximity, so the
Spanish baseline is relatively stronger for Basque specifically, leaving
less exploitable headroom. This is consistent with the headroom pattern
rather than contradicting it.

Yoruba shows reward variance collapse at the final checkpoint ($-$3.43
chrF++) with best-checkpoint selection recovering a positive gain
($+$0.11), replicating the collapse mechanism observed for Tibetan in
Experiment~B (\S\ref{sec:analysis}) and confirming that checkpoint
selection is a necessary safeguard regardless of source language.  Note that pooling Spanish and English observations to compute a single
headroom correlation is not meaningful: absolute baseline values are
source-language dependent (a Spanish baseline of 40 reflects different
NLLB-200 behaviour than an English baseline of 40), so ranks across
conditions are not directly comparable. The appropriate statistic is
the rank correlation \emph{between} conditions, reported above.

\begin{table}[h]
\centering
\footnotesize
\setlength{\tabcolsep}{4pt}
\begin{tabular}{lcccc}
\toprule
& \multicolumn{2}{c}{\textbf{English (\texttt{eng})}} &
  \multicolumn{2}{c}{\textbf{Spanish (\texttt{spa})}} \\
\cmidrule(lr){2-3} \cmidrule(lr){4-5}
\textbf{Lang.} & \textbf{Base} & \textbf{$\Delta$chrF++}
               & \textbf{Base} & \textbf{$\Delta$chrF++} \\
\midrule
\texttt{zho\_Hant} & 13.90 & $+$3.94 & 10.43 & \textbf{$+$4.05} \\
\texttt{bod\_Tibt} & 23.37 & $+$3.01 & 22.24 & $+$2.71 \\
\texttt{eus\_Latn} & 47.75 & $+$2.44 & 44.63 & $+$0.94 \\
\texttt{ben\_Beng} & 45.56 & $+$1.98 & 37.17 & $+$0.89 \\
\texttt{tur\_Latn} & 52.68 & $+$0.91 & 42.50 & $+$0.82 \\
\texttt{yor\_Latn} & 22.70 & $+$0.93 & 21.50 & $+$0.11 \\
\texttt{arb\_Arab} & 51.12 & $+$0.08 & 39.42 & $+$0.12 \\
\midrule
\multicolumn{5}{l}{\small $\rho = 0.893$, $p = 0.007$ (EN vs.\ ES rank)} \\
\bottomrule
\end{tabular}
\caption{GRPO $\Delta$chrF++ on FLORES-200 devtest (600M) for English
and Spanish source languages, sorted by English-source gain. Baselines
differ between conditions because NLLB-200 performs differently on
English$\to$X vs.\ Spanish$\to$X pairs. The rank ordering of gains is
highly consistent across source languages ($\rho = 0.893$, $p = 0.007$),
confirming that gain magnitude is driven by target-language properties
rather than source-language training dynamics.}
\label{tab:spanish_replication}
\end{table}

\section{Training Data Size Ablation}
\label{app:datasize}

Table~\ref{tab:abl_datasize} reports the full data size ablation results for
Basque (\texttt{eus\_Latn}) and Traditional Chinese (\texttt{zho\_Hant})
across $N \in \{100, 250, 500, 1{,}000\}$ training sentences. A reliable
gain threshold emerges at $N = 500$: below this, neither SFT nor GRPO
produces statistically significant improvements. At $N = 1{,}000$, GRPO
significantly outperforms SFT for Basque ($p = 0.019$) but not for
Traditional Chinese, where both methods approach the reward signal's
discriminability ceiling.

\section{Catastrophic Forgetting Results}
\label{app:forgetting}

To assess whether LoRA fine-tuning on one language degrades NLLB-200's
performance on unseen languages, each trained adapter is evaluated on three
held-out languages: French (\texttt{fra\_Latn}), Hindi (\texttt{hin\_Deva}),
and Russian (\texttt{rus\_Cyrl}). A forgetting event is defined as a chrF++
degradation exceeding 1.0 point relative to the unfine-tuned baseline on the
held-out language.

Across 39 GRPO adapter evaluations, zero forgetting events are observed.
The distribution of chrF++ deltas on held-out languages is entirely positive or
negligibly negative (range: $-$0.16 to $+$1.13, mean $+$0.34). The single
forgetting event in the entire dataset occurs in the SFT condition: a
Belarusian SFT adapter degrades Hindi by $-$0.75 chrF++. These results confirm
that the LoRA + frozen base model configuration effectively preserves
NLLB-200's other target language representations throughout GRPO training.

\section{Reward Discriminability Results}
\label{app:reward_diag}

Table~\ref{tab:reward_diagnostic} reports Pearson $r$ between reward score
and quality rank for three representative LaBSE/COMET-Kiwi weight
configurations across 12 languages, evaluated on a six-level quality cline
of 50 sentences per language from the FLORES-200 development set. The cline
consists of the human reference translation, four progressively degraded
variants, and a nonsensical candidate. All configurations achieve
good-to-excellent discrimination ($|r| > 0.81$ across all language-weight
combinations), demonstrating robustness to the exact weighting choice.

The hybrid 0.50/0.50 configuration achieves the highest mean discrimination
($-$0.930) and is optimal for most morphologically rich languages. LaBSE-only
is the best single-component configuration for Tibetan, Yoruba, Japanese, and both
Chinese variants, consistent with weaker COMET-Kiwi calibration for languages
underrepresented in its training data. For Arabic and Belarusian, the hybrid
achieves the highest discrimination of any configuration, supporting the use
of both components for root-pattern and Cyrillic fusional morphology.

\begin{center}
\footnotesize
\setlength{\tabcolsep}{4pt}
\begin{tabular}{llccc}
\toprule
\textbf{Group} & \textbf{Lang.}
  & \textbf{LaBSE} & \textbf{Hybrid} & \textbf{CK} \\
  & & \textbf{1.00/0.00} & \textbf{0.50/0.50} & \textbf{0.00/1.00} \\
\midrule
Aggl. & \texttt{eus} & $-$0.895 & \textbf{$-$0.947} & \textbf{$-$0.947} \\
      & \texttt{tur} & $-$0.888 & $-$0.940          & \textbf{$-$0.950} \\
      & \texttt{swh} & $-$0.904 & $-$0.945          & \textbf{$-$0.948} \\
      & \texttt{bod} & \textbf{$-$0.890} & $-$0.872  & $-$0.119 \\
      & \texttt{jpn} & \textbf{$-$0.880} & $-$0.864  & $-$0.836 \\
\midrule
Isol. & \texttt{zho\_Hant} & \textbf{$-$0.892} & $-$0.885 & $-$0.866 \\
      & \texttt{zho\_Hans} & \textbf{$-$0.890} & $-$0.887 & $-$0.877 \\
      & \texttt{yor}       & \textbf{$-$0.942} & $-$0.929 & $-$0.813 \\
\midrule
Root. & \texttt{arb} & $-$0.923 & \textbf{$-$0.964} & $-$0.937 \\
\midrule
Fus. & \texttt{bel} & $-$0.906 & \textbf{$-$0.958} & $-$0.940 \\
     & \texttt{ben} & $-$0.918 & \textbf{$-$0.943} & $-$0.930 \\
     & \texttt{ces} & $-$0.897 & \textbf{$-$0.955} & $-$0.923 \\
     & \texttt{pol} & $-$0.887 & \textbf{$-$0.949} & $-$0.932 \\
\midrule
\multicolumn{2}{l}{\textit{Mean (excl.\ \texttt{bod})}}
  & $-$0.902 & \textbf{$-$0.930} & $-$0.908 \\
\bottomrule
\end{tabular}
\captionof{table}{Reward discriminability: Pearson $r$ between reward
score and quality rank on a six-level quality cline (50 sentences per
language, FLORES-200 dev). \textbf{Bold} = best per language; ties
both bolded. CK = COMET-Kiwi. Tibetan (\texttt{bod}) shows near-zero
CK discriminability ($r = -0.119$); LaBSE-only strongly preferred.
Mean excludes Tibetan for comparability with \S\ref{sec:analysis}.}
\label{tab:reward_diagnostic}
\end{center}

We leave a systematic sweep over reward weights during GRPO training to future work; the present analysis establishes that the equal-weight default is a principled starting point given the components' complementary coverage and comparable discriminative power.

\section{Decoding Regime Control Ablation}
To assess whether GRPO gains reflect genuine policy improvement rather
than decoding regime differences, we evaluate the unfine-tuned 600M
baseline under temperature sampling ($T = 1.2$, matching GRPO training) on four languages. Temperature sampling of the untrained
model substantially \emph{degrades} chrF++ relative to beam search (Basque: $47.64 \to 30.03$; Arabic: $51.03 \to 27.86$). GRPO trained under the same sampling regime achieves $+$2.93 chrF++ over the beam search baseline for Basque and $+5.07$ for Traditional Chinese, substantially exceeding any decoding regime effect. The decoding regime therefore cannot account for the observed gains.

\section{Human Preference Evaluation}
\label{app:human_eval}

\subsection{Setup}

We evaluate four languages spanning a range of GRPO gain magnitudes:
Traditional Chinese (\texttt{zho\_Hant}, $+$3.94 chrF++), Bengali
(\texttt{ben\_Beng}, $+$1.98), Turkish (\texttt{tur\_Latn}, $+$0.91),
and Polish (\texttt{pol\_Latn}, $+$0.41). These languages were selected
to cover both the upper and lower ends of the gain distribution,
allowing us to characterize not just whether gains are perceptible
but at what magnitude they become so.

For each language, 50 source sentences were sampled uniformly from
the FLORES-200 devtest set. Each sentence was paired with two
translations: the Baseline NLLB-200 output (beam search, no fine-tuning)
and the GRPO output. Annotators were presented with
the source sentence and both translations in randomised order, with
no indication of which system produced each output. They were asked
to select the translation they considered higher quality, or to mark
the pair as a tie if both were equivalent. No time limit was imposed.
Annotation was conducted by bilingual speakers with native or
near-native proficiency in the target language. Each sentence pair
was evaluated by a single annotator.

\subsection{Statistical Analysis}

Preference counts were analysed using a one-sided binomial test
against the chance hypothesis (equal preference for GRPO and Baseline
among non-tied pairs). Ties were excluded from the binomial test
following standard practice in MT preference evaluation
\citep{graham2015accurate}. Sentences with missing annotations
(annotator dropout) were excluded from all counts; effective sample
sizes are reported in Table~\ref{tab:human_eval}.

\subsection{Results}

\begin{table}[h]
\centering
\footnotesize
\setlength{\tabcolsep}{4pt}
\begin{tabular}{lccccc}
\toprule
\textbf{Lang.} & \textbf{$\Delta$chrF++}
  & \textbf{GRPO\%} & \textbf{Base\%} & \textbf{Tie\%} & \textbf{$p$} \\
\midrule
\texttt{zho\_Hant} & $+$3.94 & 68.0 &  8.0 & 24.0 & $<$0.001\\
\texttt{ben\_Beng} & $+$1.98 & 20.0 & 31.1 & 48.9 & 0.895\\
\texttt{tur\_Latn} & $+$0.91 & 42.0 & 42.0 & 16.0 & 0.561\\
\texttt{pol\_Latn} & $+$0.41 & 34.8 & 41.3 & 23.9 & 0.750\\
\bottomrule
\end{tabular}
\caption{Human preference results (50 sentences per language,
45--50 after excluding missing annotations). $p$: one-sided
binomial test on non-tied pairs. $\Delta$chrF++ from 600M, Experiment~A.}
\label{tab:human_eval}
\end{table}

Results reveal a clear perceptibility threshold. For Traditional
Chinese, where GRPO achieves the largest automatic gain ($+$3.94
chrF++), annotators strongly and significantly prefer GRPO output
(68\% vs.\ 8\% Baseline, $p < 0.001$). For the three remaining
languages --- Bengali, Turkish, and Polish, with gains of $+$1.98,
$+$0.91, and $+$0.41 chrF++ respectively --- preferences are
indistinguishable from chance (all $p > 0.5$). This pattern is
consistent with established findings that sub-2-point chrF++
improvements are typically below the threshold of human
perceptibility \citep{graham2015accurate}.

Two aspects of the small-gain results deserve attention. First,
Turkish and Polish show near-exact parity between GRPO and Baseline
preference (42\%/42\% and 35\%/41\%) rather than a Baseline
advantage, indicating the policy does not degrade perceived
translation quality even when automatic metric gains are small.
Second, the Bengali result --- where Baseline is slightly preferred
despite a $+$1.98 chrF++ gain --- suggests that for moderate gains,
GRPO may introduce changes that automatic metrics reward but that
do not correspond to human-perceptible improvements; this is
consistent with the known sensitivity of COMET-family metrics to
surface fluency features that human judges do not weight equally.

\subsection{Scope and Limitations}

This evaluation is intentionally limited in scale. Each language
was assessed by a single annotator on 50 sentences, with no
inter-annotator agreement measurement. The goal was not to
produce a definitive human evaluation but to perform a targeted
sanity check: do the largest automatic gains correspond to
perceptible improvements? For Traditional Chinese the answer is
unambiguously yes. For smaller gains, the evaluation is
underpowered to draw strong conclusions, and a larger study
with multiple annotators and inter-annotator agreement analysis
would be needed to characterise the perceptibility threshold
more precisely. We present these results as a frugal validation
exercise rather than a substitute for full human evaluation,
and note that the direction and pattern of results are consistent
with what the headroom analysis would predict: gains large enough
to matter are large enough to be noticed.

\end{document}